\pdfoutput=1

\documentclass[11pt]{article}

\usepackage{emnlp2021}

\usepackage{times}
\usepackage{latexsym}

\usepackage[T1]{fontenc}

\usepackage[utf8]{inputenc}

\usepackage{microtype}
\usepackage{graphicx}
\title{Finetuning Pretrained Transformers into Variational Autoencoders}

\author{Seongmin Park\qquad Jihwa Lee \\
    ActionPower\\
  Seoul, Republic of Korea \\
  \texttt{\{seongmin.park, jihwa.lee\}@actionpower.kr} \\}

\begin{document}
\maketitle
\begin{abstract}
Text variational autoencoders (VAEs) are notorious for \textit{posterior collapse}, a phenomenon where the model's decoder learns to ignore signals from the encoder. Because posterior collapse is known to be exacerbated by expressive decoders, Transformers have seen limited adoption as components of text VAEs. Existing studies that incorporate Transformers into text VAEs \cite{li2020optimus, fang2021transformer} mitigate posterior collapse using massive pretraining, a technique unavailable to most of the research community without extensive computing resources. We present a simple two-phase training scheme to convert a sequence-to-sequence Transformer into a VAE with just \textit{finetuning}. The resulting language model is competitive with massively pretrained Transformer-based VAEs in some internal metrics while falling short on others. To facilitate training we comprehensively explore the impact of common posterior collapse alleviation techniques in the literature. We release our code for reproducability\footnote{\url{https://github.com/seongminp/transformers-into-vaes}}.
\end{abstract}

\section{Introduction}

Properly tamed latent models offer explainable and interpolatable representations of observed data. Recent works have shown such models to be especially useful in unsupervised learning settings. \citet{he2019probabilistic} adapt a generative latent text model for successful unsupervised text style transfer and machine translation. \citet{li2020optimus} achieve superior language modeling performance against common conditional counterparts. 

A popular variant of deep latent models is the variational autoencoder (VAE) \citep{Kingma2014}. For each observed \textbf{x}, the model assumes the existence of a corresponding multi-dimensional latent vector \textbf{z}. Since the log evidence $\log{p(x)}$ is intractable for most interesting problems, the training process for VAEs opts instead to minimize the log evidence lower bound (ELBO): 
\begin{equation}
E_{z\sim q(z|x)}[log(p(x|z))] - D_{KL}(q(z|x)||p(z))
\end{equation}
$q(z|x)$ is a tractable, assumed posterior commonly modeled with a parametrized encoder $q_{\phi}(z|x)$, while $p(x|z)$ is the likelihood parametrized with a decoder $p_{\theta}(x|z)$ that optimizes against reconstruction loss. While effective in theory, a common empirical challenge VAEs present during training is posterior collapse – a phenomenon where the decoder ignores the latent signal from \textbf{z} (and thus the originating input) during reconstruction. Posterior collapse can be diagnosed by checking if $D_{KL}(q(z|x) || p(z))$ tends to zero during training. 

After \citet{bowman2016generating} adopted VAE for text, subsequent studies have been introduced with attempts to mitigate posterior collapse in VAE language models (LMs). However, the brittle training process of VAE LMs remains an unsolved problem.  

\citet{li2020optimus} present a method to utilize deep Transformer \citep{NIPS2017_3f5ee243} models as components of VAE LMs. Transformer-based VAEs tap into the state-of-the-art capabilities of Transformers while retaining representational advantages of VAE LMs. The paper mitigates posterior collapse by massive pretraining and a cyclical annealing schedule \citep{fu2019cyclical}.

While the study presents a promising outlook for Transformer VAEs, the suggested method is not accessible to researchers who lack access to large, target-domain-specific corpora or the computing power for massive LM pretraining. Therefore, a demand arises for a way to finetune an existing Transformer model into a VAE LM with limited resources. Our research attempts to fill this gap in the literature, and makes the following contributions: 

\begin{itemize}
  \item We present a simple but reliable (as replicated across several datasets) scheme to teach latent structure to a pretrained Transformer model by just finetuning.
  \item We convert a pretrained sequence-to-sequence Transformer into a VAE, instead of using two separate encoder-only \citep{devlin-etal-2019-bert} and decoder-only \citep{radford2019language} Transformers as in previous literature. This eliminates the need to maintain separate tokenizers and configurations for encoder and decoder. 
  \item We conduct ablation studies and extensive experiments to gauge the effectiveness of commonly used posterior collapse mitigation methods in taming Transformer VAEs. 

\end{itemize}

The resulting model extends existing Transformer architectures and can be initialized from pretrained non-latent model checkpoints. 

\section{Background}

\subsection{Transformer Text VAEs}

Most VAE LMs employ recurrent neural networks (RNNs) as encoders and decoders. This is in part because enforcing a latent bottleneck layer undermines the effectiveness of encoder-decoder cross-attention in Transformers, and in significant part due to the co-occurrence of posterior collapse and powerful and deep decoder layers. \citet{li2020optimus} overcome such training difficulties by massively increasing the number of training samples (104,213,036 sentences) for LM pretraining. 

\citet{liu2019transformer} and \citet{fang2021transformer} also finds success with Transformer VAEs for text generation. To avoid posterior collapse, \citet{fang2021transformer} follow the exact cyclic KL mitigation approach as that of \citet{li2020optimus}, while \citet{liu2019transformer} introduce noise to network input.  

\subsection{Techniques to mitigate posterior collapse}
This study identifies and explores the effect of popular posterior collapse mitigation methods in low-resource Transformer VAE training. We do not examine importance-weighted autoencoders \cite{burda2016importance} and semi-amortized autoencoders \cite{pmlr-v80-kim18e} to limit the scope of our experiments to unsophisticated prior distributions.

\subsubsection{KL Weighting / Annealing}
\citet{bowman2016generating} increases the KL term of the ELBO from zero to its full value during early stages of training, where the decoder learns to simply treat latent signal \textbf{z} as noise. \citet{fu2019cyclical} extend this technique by cyclically manipulating the weight of the KL term. $\beta$-VAE \citep{DBLP:conf/iclr/HigginsMPBGBML17} and \citet{10.1145/3388440.3412458} adopt a similar approach. 
\subsubsection{Encoder warm-up \cite{li-etal-2019-surprisingly}}
We train the network without the KL term of the ELBO and retain encoder weights before jointly training the whole network. 

\subsubsection{Input text denoising \cite{shen2020educating}}
Denoising text inputs by deleting random tokens motivate autoencoders (AEs) to learn better latent representations. Our study compares 0\%, 15\%, and 40\% deletion noising schemes.     
\subsubsection{KL thresholding \cite{NIPS2016_ddeebdee}} 
KL-thresholding enforces a minimum $\lambda$ for each dimension of the KL term in the ELBO: 
\begin{equation}
\mathcal{L}_{D_{KL}} = \sum_{i} max[\lambda, D_{kl}(q_\phi(z_i|x)||p(z_i))]
\end{equation}
where $z_i$ is a single dimension of \textbf{z}.

\subsubsection{Encoder pooling \cite{long2019preventing}}
Instead of using the last hidden state as encoder output, averaging or taking the maximum of all encoder hidden states results in a more diverse latent representation. We experiment with both mean- and max-pooling schemes from the encoder. 

\section{Model architecture}

\begin{figure*}[h]
\includegraphics[height=5cm,keepaspectratio]{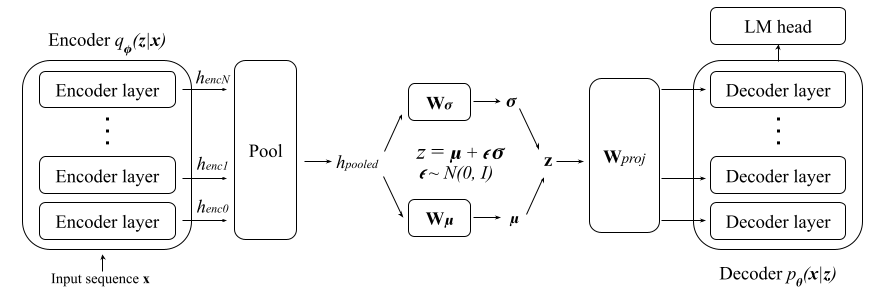}
    \caption{Transformer VAE architecture. A "bottleneck" step ($W_{\sigma}$ and $W_{\mu}$) is placed between the encoder and the decoder of T5. Latent information from pooled encoder hidden states is captured in the bottleneck layer before being passed to the decoder. The network is optimized against regularization loss in the bottleneck and reconstruction loss at the decoder.}
\label{fig:architecture}
\end{figure*}

We extend the T5 architecture \cite{JMLR:v21:20-074} into a VAE. We modify a popular pretrained T5 model \cite{wolf-etal-2020-transformers} that deviates minimally from the original Transformer (Figure \ref{fig:architecture}).

Hidden states from all layers of T5’s encoder $q_{\phi}(z|x)$ are mean- or max-pooled into a vector $h_{pooled} \in R^{H}$, where $H$ is the encoder's hidden dimension.  

Assumed prior $q(z)$’s mean $\mu$ and log variance $\sigma$ vectors of dimension $L$ is obtained from $h_{pooled}$: 
\begin{equation}
\mu = h_{pooled} W_{\mu},\ log \sigma = h_{pooled} W_{\sigma}
\end{equation}

where $W_{\mu}, W_{\sigma} \in R^{L}$.

As in a standard VAE, a stochastic latent vector \textbf{z} is sampled using the reparameterization trick to enable back-propagation through sampling: 

\begin{equation}
z = \mu + \sigma\odot\epsilon,\ \epsilon\sim \mathcal{N}(0, 1)
\end{equation}

We pass \textbf{z} into the decoder $p_{\theta}(x|z)$ as the only component of decoder’s cross attention. Our method injects \textbf{z} into every layer of the decoder as in previous literature \cite{li2020optimus,fang2021transformer}, but deviates in two important ways: first, we pass \textbf{z} as the sole key and value of encoder-decoder cross attention, instead of self-attention; second, we project z into the correct dimension ($L\times A\times S$, where $L$ is the decoder layer count, $A$ is the number of attention heads, and $S$ is the embedding dimension per head) with a feed-forward network, instead of taking a copy of \textbf{z} to inject to each decoder layer.  
\begin{equation}
(K_{ca}, V_{ca}) = (zW_{proj}, zW_{proj})
\end{equation}

where $W_{proj} \in R^{L\times A\times S}$ and $K_{ca}$ and $V_{ca}$ are key and value in decoder cross-attention.

\section{Experiments}

During preliminary experiments, posterior collapse was observed in all training schemes without encoder warmup training. The decoder learns to ignore the initially noisy input signal from the encoder. Thus, we compose our finetuning method in two separate phases.

\textbf{Phase 1 - Encoder warmup}: Weight of KL loss is set to zero, making our model’s objective function similar to that of an AE. Different input denoising percentages, encoder pooling strategies, latent dimension sizes, and decoder freezing configurations are compared. 

\textbf{Phase 2 - Full finetuning}: KL loss is reinstated and full VAE training are conducted. We compare different input denoising percentages, encoder pooling strategies, KL annealing schedules, and KL thresholds. 

We run our proposed two-phase finetuning training scheme on four standard VAE LM benchmark datasets: PTB \cite{marcus-etal-1993-building}, SNLI \cite{bowman2016generating}, Yahoo \cite{10.5555/3305890.3306082}, and Yelp \cite{10.5555/3295222.3295427}.  

Following \citet{li2020optimus} and \citet{li-etal-2019-surprisingly}, we perform intrinsic evaluation of our proposed Transformer VAE architecture. We report perplexity (PPL), KL-divergence between model posterior and assumed posterior (KL), and negative ELBO on the test set. To assess the quality of learned latent codes, we also report mutual information (MI) \cite{hoffman2016elbo} and the number of active units (AU) \cite{burda2016importance}. MI measures the dependence of latent codes to encoder input. AU measures the covariance between encoder input and latent codes.

Experimental hyperparameters such as specific annealing schedules and training epochs per phase are detailed in the appendix.

\section{Results}

\begin{table*}[ht!]
\centering
\begin{tabular}{l|lllll}
\hline
    \textbf{Model} & \textbf{PPL}$\downarrow$ & \textbf{KL} & \textbf{-ELBO}$\downarrow$  & \textbf{MI}$\uparrow$ & \textbf{AU}$\uparrow$\\
\hline
    Optimus ($\lambda$ = 0.5) \cite{li2020optimus} & 23.11 & 17.45 & 301.21 & 8.85 & 32\\
    GPT-2 \cite{radford2019language}\ & 22.00 & - &- & - & -\\
    Encoder pretraining ($\lambda$ = 3) \cite{li-etal-2019-surprisingly}\ & 59.24 & 7.44 & 328.73 & 6.41 & 32 \\
\hline
Ours (Max pool) & 20.90 & 0.21 & 343.02 & 0.04 & 0\\
Ours (Max pool + Denoise)& 30.13 & 41.49 & 301.86 & 1.32 & 24\\
Ours (Max pool + Denoise + KLT)& 60.44 & 119.89 & 223.69 & 4.73 & 29 \\
Ours (Max pool + Denoise + KLT + Deep)& 54.40 & 155.50 & 140.57 & 5.43 & 28 \\
\hline
\end{tabular}
\label{results}
\caption{Phase 2 results on Yahoo. Due to space constraints, we report experimental results on other datasets in the appendix. Results on baselines are quoted from \citet{li2020optimus} and \cite{li-etal-2019-surprisingly}. KLT denotes KL thresholding with $\lambda=3$. Our models are finetuned from a pretrained 6-layer T5, except the deep variant with 12 layers.}
\end{table*}

\subsection{Phase 1} 

\begin{figure}[h]
\includegraphics[width=\columnwidth]{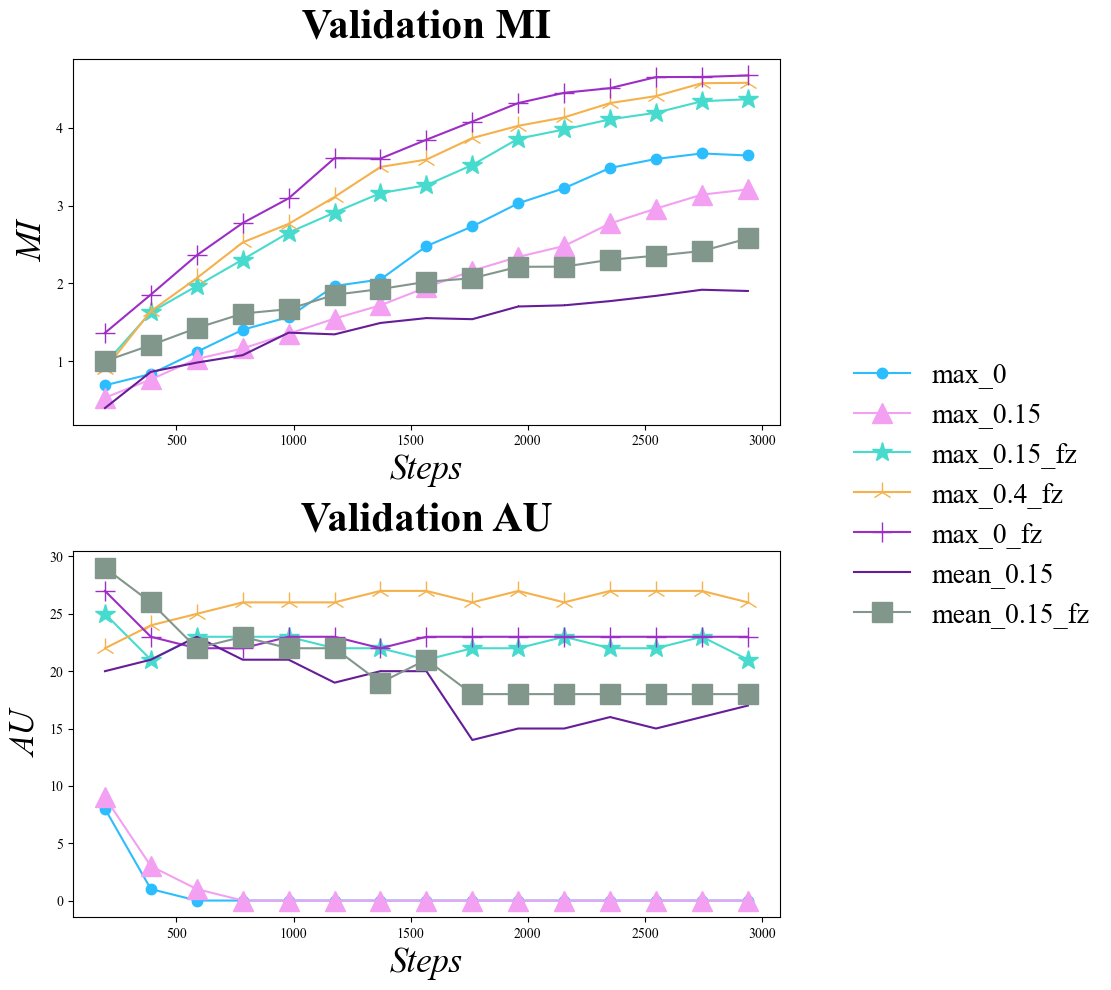}
\caption{Phase 1 training on Yahoo. Labels are in the form \{\textit{pooling strategy}\}\_\{\textit{denoise percentage}\}\_\{\textit{decoder frozen}\}.}
\label{fig:phase1}
\end{figure}

We find that freezing the decoder and the memory projection layer $W_{proj}$ while training with an AE objective is crucial in learning meaningful encoder outputs. 
Denoising is important for datasets with longer inputs (Yahoo, Yelp), but not critical in datasets with shorter input lengths (PTB, SNLI). Mean-pooling encoder hidden states presents a trade-off between MI and AU. Max-pooling consistently learns more informative encoder representations. Changes in MI and AU during training is illustrated in Figure \ref{fig:phase1}. 

Latent dimensions of 64 and 128 were also tested. Increasing the latent dimension did not necessarily boost representational quality in terms of AU percentage. For latent dimensions of 32 and 64, 90\% of latent dimension units were activated in best-performing models. For latent dimension of 128, around 60\% of latent units were active.   

Another interesting observation is that KL divergence on the validation set, although not part of the AE training objective, plateaus after repeated training. We regard this phenomenon as the signal of convergence in terms of representation quality.

\subsection{Phase 2}

We observe, as in previous literature, a trade-off between language modeling PPL and representation quality metrics (MI and AU). This trade-off is exacerbated when using KL thresholding. While KL thresholding does significantly increase latent representation capabilities, it is not in itself sufficient in preventing posterior collapse.

\begin{figure}[h]
\includegraphics[width=\columnwidth]{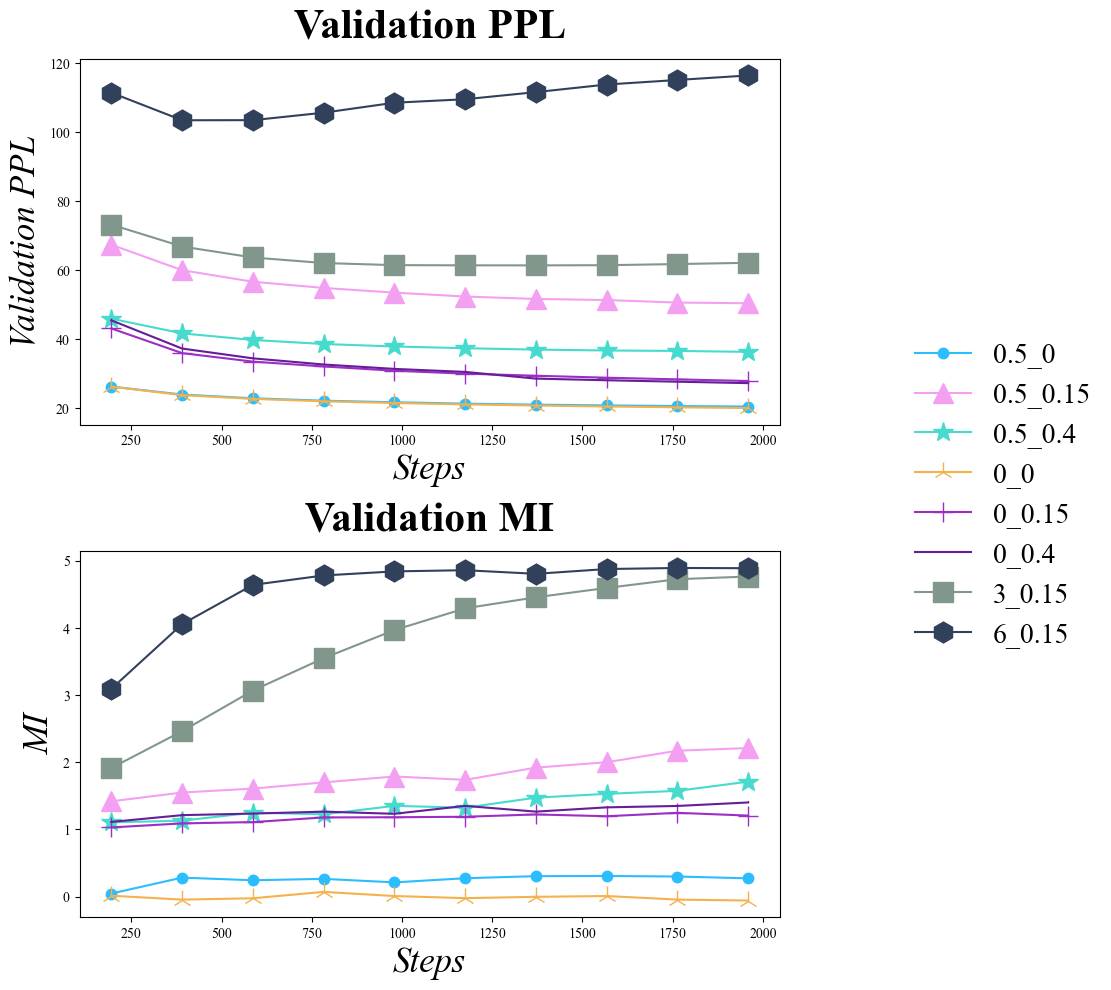}
\caption{Phase 2 training on Yahoo. Labels are in the form \{\textit{KL threshold}\}\_\{\textit{denoise percentage}\}. Encoder hidden states in plotted experiments were max-pooled.}
\label{fig:phase2}
\end{figure}

Denoising and encoder pooling configurations display the same characteristics as in Phase 1. No version of the experiment existed where cyclical annealing schedule was able to prevent posterior collapse, a result not in accordance with \citet{li2020optimus}.  Figure \ref{fig:phase2} illustrates the training progression of Phase 2. 

We also experimented with increasing model depth from 6 layers to 12 layers. Our proposed two-phase training scheme prevents posterior collapse for deeper models as well, resulting in higher performance in most metrics compared to 6-layer models. Results are reported in Table 1. Note that lower PPL does not necessarily indicate better language modeling capabilities, since models with collapsed posterior display better PPL. 

Rows with KL above zero indicate successful aversion of posterior collapse.

In the literature, no consensus yet exists on the optimal value of KL in training VAEs. Overall, we find that a denoising scheme between 0.15 and 0.4 in both phases, coupled with a low (0.5) KL threshold strikes a good balance between reconstruction and latent representation quality. 

\section{Conclusions and Future Work}

This paper explores common methods in the literature for combatting posterior collapse, and the extent to which they help in teaching latent information to pretrained Transformer models.

Comprehensive experiments show that commonly employed posterior collapse mitigation techniques provide meaningful benefits in transforming existing language models into latent-aware architectures. Among the tested procedures, we find that \citet{li-etal-2019-surprisingly}'s two-step training, coupled with \citet{shen2020educating}'s denoising through token deletion, was the most impactful in mitigating posterior collapse. However, language models obtained via only finetuning exhibit consistent trade-offs between their latent representation metrics (MI, AU) and language model metrics (PPL). Optimizing our model to be competitive with massively pretrained baselines in one of the two metrics results in the model falling behind in the other. 

We also find that increasing training epochs further improves the impact of tested techniques, a result consistent with previous literature on large-scale text VAE pretraining.

From our experiments, we identify several questions to be answered by future research. The impact of homogenizing finetuning (as suggested in this paper) and original pretraining objectives on language model metrics has to be further explored. While the original T5 architecture was also pretrained with a self-supervised denoising scheme, the model employs mask tokens for denoising, contrary to simple token deletions suggested by this paper. 

Our findings also highlight the need for an established heuristic to interpret the quality of latent representations learned by language models. The research community has yet to decide on the optimal value of KL-divergence between the assumed prior and the model posterior to target during text VAE training. Empirical guidelines to dictate even a vague threshold for the KL-divergence, below which we declare the occurrence of posterior collapse, will help both training and evaluation of latent-aware language models.

\bibliography{anthology,custom}
\bibliographystyle{acl_natbib}
\clearpage

\appendix

\section{Phase 2 results on PTB, Yelp, and SNLI}
\label{sec:appendix}

\begin{table}[h]
\centering
\begin{tabular*}{\textwidth}{l|lllll}
\hline
\textbf{Model} & \textbf{PPL}$\downarrow$ & \textbf{KL} & \textbf{-ELBO}$\downarrow$  & \textbf{MI}$\uparrow$ & \textbf{AU}$\uparrow$\\
\hline
Optimus ($\lambda$ = 0.5) \cite{li2020optimus} & 26.69 & 15.72 & 96.82 & 7.64 & 32\\GPT-2 \cite{radford2019language}\ & 24.23 & - & - & - & -\\
Encoder pretraining ($\lambda$ = 3) \cite{li-etal-2019-surprisingly}\ & 96.75 & 3.85 & 101.56 & 3.19 & 32 \\
\hline
Ours (Max pool) & 51.60 & 0.09 & 104.14 & 0 & 0 \\
Ours (Max pool + Denoise)& 57.69 & 3.05 & 101.17 & 11& \\
Ours (Max pool + Denoise + KLT)& 250.44 & 41.93 & 62.25 & 2.17 & 24 \\
Ours (Max pool + Denoise + KLT + Deep)& 705.73  & 84.28 & 4.95 & 13 & \\
\hline
\end{tabular*}
\label{results}
\caption{Phase 2 results on PTB}
\end{table}

\begin{table}[h]
\centering
\begin{tabular*}{\textwidth}{l|lllll}
\hline
\textbf{Model} & \textbf{PPL}$\downarrow$ & \textbf{KL} & \textbf{-ELBO}$\downarrow$  & \textbf{MI}$\uparrow$ & \textbf{AU}$\uparrow$\\
\hline
Optimus ($\lambda$ = 0.5) \cite{li2020optimus} & 22.79 & 15.09 & 344.10 & 9.13 & 32\\
GPT-2 \cite{radford2019language}\ & 23.40 & - & - & - & -\\
Encoder pretraining ($\lambda$ = 3) \cite{li-etal-2019-surprisingly}\ & - & - & - & - & - \\
\hline
Ours (Max pool) & 21.65 & 0.25 & 404.54 & 0 & 0\\
Ours (Max pool + Denoise)& 39.09 & 77.85 & 327.17 & 1.06 & 26\\
Ours (Max pool + Denoise + KLT)& 86.71 & 182.24 & 223.34 & 5.46 & 27 \\
Ours (Max pool + Denoise + KLT + Deep)& 53.05 & 178.48& 166.15 & 5.55 & 10 \\
\hline
\end{tabular*}
\label{results}
\caption{Phase 2 results on Yelp.}
\end{table}

\begin{table}[h]
\centering
\begin{tabular*}{\textwidth}{l|lllll}
\hline
\textbf{Model} & \textbf{PPL}$\downarrow$ & \textbf{KL} & \textbf{-ELBO}$\downarrow$  & \textbf{MI}$\uparrow$ & \textbf{AU}$\uparrow$\\
\hline
Optimus ($\lambda$ = 0.5) \cite{li2020optimus} & 16.67 & 16.35 & 38.50 & 8.89 & 32\\
GPT-2 \cite{radford2019language}\ & 20.24 & - & - & - & -\\
Encoder pretraining ($\lambda$ = 3) \cite{li-etal-2019-surprisingly}\ & 21.23 & 5.86 & 33.87 & 5.25 & 32 \\
\hline
Ours (Max pool) & 12.79 & 0.11 & 34.61 & 0.06 & 0\\
Ours (Max pool + Denoise)& 15.16 & 2.41 & 32.32 & 0.194 & 7\\
Ours (Max pool + Denoise + KLT)& 85.88 & 26.12 & 8.62 & 0.88 & 13 \\
Ours (Max pool + Denoise + KLT + Deep)& 2358.31 & 74.95 & 43.77 & 5.27 & 17 \\
\hline
\end{tabular*}
\label{results}
\caption{Phase 2 results on SNLI.}
\end{table}

\section{Experimental details}
For all experiments we used a AdamW optimizer \cite{loshchilov2018decoupled} with a starting learning rate of $1\times10^{-3}$, $\beta_1=0.9$, $\beta_2=0.999$, and $\epsilon=1\times10^{-3}$. The linear KL annealing schedule we used was as follows:
\begin{equation}
KL\ weight = \frac{current\ global \ step}{steps\ per\ epoch * 50}
\end{equation}
Our slower, linear KL annealing schedule of 0 to 1 over 50 epochs yielded better empirical results than the linear schdule used in \citet{li-etal-2019-surprisingly} (0 to 1 over 10 epochs). We attribute this result to the\\ \\ \\ \\ \\ \\ \\ \\ \\ \\ \\ \\ \\ \\ \\ \\ \\ \\ \\ \\ \\ \\ \\ \\ \\ \\ \\ \\ \\ \\ \\ \\ \\ \\ \\ \\ \\ \\ \\ \\
small number of training samples in our experiments.

We train for 5 epochs on Phase 1, and 3 epochs on Phase 2. While further training leads to increased MI and AU, we limit the number of epochs to confer to the spirit of this study, which is to learn latent representations with minimal training. The 5 epoch limit on Phase 1 was empirically determined as the point where encoder MI begins to plateau. Most experiments were conducted with \textbf{z} dimension of 32 for comparison with previous literature.

\end{document}